\documentclass[runningheads]{llncs}

\usepackage[utf8]{inputenc}
\usepackage[T1]{fontenc}
\usepackage{amsmath, amssymb}
\usepackage{graphicx}
\usepackage{booktabs}
\usepackage{algorithm}
\usepackage{algorithmic}
\usepackage{xcolor}
\usepackage{float}
\usepackage{hyperref}

\usepackage{tikz}
\usetikzlibrary{positioning, arrows.meta, shapes.geometric, calc}

\title{\textbf{GS-KAN: Parameter-Efficient Kolmogorov-Arnold Networks via Sprecher-Type Shared Basis Functions}}

\author{Oscar Eliasson
}
\institute{Chalmers University of Technology}
\date{\today}

\begin{document}

\maketitle
\begin{abstract}
The Kolmogorov-Arnold representation theorem offers a theoretical alternative to Multi-Layer Perceptrons (MLPs) by placing learnable univariate functions on edges rather than nodes. While recent implementations such as Kolmogorov-Arnold Networks (KANs) demonstrate high approximation capabilities, they suffer from significant parameter inefficiency due to their reliance on unique parameterizations for every network edge. In this work, we propose GS-KAN (Generalized Sprecher-KAN), a lightweight architecture inspired by David Sprecher's refinement of the superposition theorem. GS-KAN constructs unique edge functions by applying learnable linear transformations to a single learnable, shared parent function per layer. We evaluate GS-KAN against existing KAN variants and MLPs across synthetic function approximation, real-world regression, and image classification tasks. Our experimental results demonstrate that GS-KAN achieves the strongest predictive performance among all evaluated baselines across the three tested domains. Crucially, the proposed architecture enables the deployment of KAN-based architectures in high-dimensional regimes under strict parameter constraints, a setting where standard implementations are typically infeasible due to parameter explosion. The source code is available at \url{https://github.com/rambamn48/gs-impl}.
\end{abstract}

\section{Introduction}
The Multi-Layer Perceptron (MLP) has long served as the foundational building block of deep learning. By stacking linear transformations followed by fixed element-wise non-linearities (e.g., ReLU), MLPs leverage the Universal Approximation Theorem \cite{cybenko1989,hornik1991} to model complex data manifolds. However, the recent introduction of Kolmogorov-Arnold Networks (KANs) \cite{liu2024kan} challenges this paradigm by placing learnable univariate functions on the edges of the network. KANs have shown promise in interpretability and data efficiency, particularly for low-dimensional scientific discovery tasks.

However, standard KANs face a critical scalability bottleneck. The reliance on unique function parameterization for every edge leads to a parameter complexity of $O(C \cdot N_{in} \cdot N_{out})$ per layer, where $C$ is the number of parameters for the learned functions. For high-dimensional inputs, such as flattened images, this results in an explosion of parameters that renders standard KANs impractical or prohibitive to deploy under strict memory budgets. While recent variants like Wav-KAN \cite{bozorgasl2024wavkan} attempt to address this by substituting splines with fixed wavelets, they often sacrifice the learnable adaptability of the basis function.

In this paper, we revisit the theoretical foundations laid by Sprecher \cite{sprecher1965} to address this inefficiency. While the original KAN architecture \cite{liu2024kan} leverages the Kolmogorov-Arnold representation theorem \cite{kolmogorov1963}, it does not fully exploit Sprecher's subsequent refinement, which demonstrated that the inner functions can be constructed as transformed versions of a \textit{single} univariate function. Historically, direct implementations of such ``Sprecher networks'' were hindered by the pathological, fractal nature of the exact functions required by the theorem, rendering them unsuitable for gradient descent (see, e.g., \cite{demb2021note}).

We propose GS-KAN (Generalized Sprecher-KAN), which relaxes these strict conditions. Instead of attempting to model exact fractal functions, we employ smooth, learnable B-splines combined with generalized coefficients. Unlike standard KANs which maintain unique splines for every edge, GS-KAN maintains one learnable master function per layer. This reduces the parameter complexity, effectively aligning it with MLP efficiency while retaining the powerful inductive bias of splines. Our contributions are as follows:
\begin{itemize}
    \item \textbf{Parameter Efficiency:} We propose a memory-efficient formulation using layer-wise shared learnable B-splines, adapted via learnable linear parameters. This architecture significantly reduces the parameter footprint compared to standard KANs.
    \item \textbf{Scalability:} By decoupling the basis definition from the network edges, GS-KAN enables the use of spline-based architectures on high-dimensional inputs (e.g., image vectors), overcoming the parameter bottlenecks of previous implementations.
    \item \textbf{Empirical Validation:} Our results indicate that GS-KAN is a versatile architecture capable of effectively handling diverse machine learning tasks under strict parameter constraints. Notably, it achieved the strongest predictive performance among all evaluated baselines across our entire test suite.
\end{itemize}

A recent concurrent study by \cite{hagg2025sprecher} also explores the application of Sprecher’s theorem for parameter efficiency. Their approach focuses on a strict realization of Sprecher's specific summation formula using shared weight vectors. In contrast, our work builds upon a variation of the theorem and from this basis, we propose a generalized architecture where we relax the fixed constant constraints to construct unique edge functions via learnable per-edge linear transformations. This distinction allows our method to retain the memory benefits of the theorem while adopting the flexible connectivity typical of modern deep learning layers.

This work extends the initial investigations documented in the author's Bachelor's thesis \cite{eliasson2025kolmogorov}. Although the core concept of leveraging Sprecher's theorem originated there, the present work introduces a fully generalized architecture explicitly designed to support arbitrary input-output mappings—including multi-output regression and classification tasks—accompanied by improved initialization strategies and deeper empirical analysis.

\section{Theoretical Background}

This section outlines the mathematical framework underpinning neural networks. By composing elementary operations into layers and stacking them sequentially, we establish the deep architecture required for complex modeling. The following theorems describe the capabilities of this hierarchical structure.

\subsection{Universal Approximation Theorem (MLP)}

The Multi-Layer Perceptron is grounded in the Universal Approximation Theorem, initially formulated by \cite{cybenko1989} and \cite{hornik1991}. This theorem guarantees that MLPs can act as universal function approximators under general conditions.

\begin{theorem}[Universal Approximation Theorem]
\label{thm:uat}
Let $\sigma(\cdot)$ be a fixed, non-linear activation function. Any continuous function $f: \mathbb{R}^n \to \mathbb{R}$ can be approximated to arbitrary accuracy by a finite linear combination of the form:
\begin{equation}
    f(\mathbf{x}) \approx \sum_{i=1}^{N} v_i \sigma \left( \sum_{j=1}^{n} w_{ij} x_j + b_i \right)
\end{equation}
where $N$ is the number of hidden neurons, and $w_{ij}, v_i, b_i$ are learnable parameters.
\end{theorem}

\subsection{Kolmogorov-Arnold Representation (KAN)}
In contrast to MLPs, Kolmogorov \cite{kolmogorov1963} proved that any continuous multivariate function on a compact domain $[0,1]^n$ can be represented exactly as a superposition of continuous univariate functions. The proposed {KAN} architecture \cite{liu2024kan} parameterizes these univariate functions $\phi_{q,p}$ as B-splines.

\begin{theorem}[Kolmogorov-Arnold Representation]
For any continuous function $f: [0,1]^n \to \mathbb{R}$, there exist continuous univariate functions $\Phi_q$ and $\phi_{q,p}$ such that:
\begin{equation}
    f(x_1, \dots, x_n) = \sum_{q=0}^{2n} \Phi_q \left( \sum_{p=1}^{n} \phi_{q,p}(x_p) \right)
\end{equation}
\end{theorem}

\subsection{Sprecher's Refinement}
Sprecher \cite{sprecher1965} refined the representation theory by demonstrating that distinct functions for each branch are not required. Instead, the theorem guarantees that a multivariate function can be represented using \textit{single} continuous functions $\chi$ and $\psi$ through a specific construction involving scaling and translation. Here, $\psi$ is a universal function shared across all terms. Crucially, the outer function $\chi$ is also {shared} (independent of $q$), acting on the aggregated summation.

\begin{theorem}[Sprecher, 1965]
For any continuous function $f: [0,1]^n \to \mathbb{R}$, there exist real constants $\lambda, \epsilon$ and a continuous function $\chi: \mathbb{R} \to \mathbb{R}$ such that:
\begin{equation}
    f(\mathbf{x}) = \sum_{q=0}^{2n} \chi \left( \sum_{p=1}^{n} \lambda^{p\cdot q} \psi(x_p + \epsilon q) \right)
\end{equation}

\end{theorem}

\subsection{Related Efficient Architectures}
Recent works have attempted to streamline KANs. Notably, {Wav-KAN} \cite{bozorgasl2024wavkan} replaces B-splines with wavelets, defined as $\psi_{w,s,t}(x) = w \psi(\frac{x-t}{s})$. While Wav-KAN shares the philosophy of transforming a basis function, it typically relies on a fixed analytical mother wavelet (e.g., Mexican Hat). Our approach differs by making the basis function $\psi$ itself a learnable B-spline, optimizing both the transformation and the shape of the function simultaneously.


\section{Methodology}

To translate Sprecher's theoretical framework into a flexible deep learning network, we introduce key relaxations to the strict formulation. While the theorem postulates the existence of specific fixed constants to ensure exact representation, enforcing such rigid values restricts the model's adaptability for gradient-based optimization.

\subsection{Generalized Representation}
Our architecture, {GS-KAN}, relaxes the theorem's fixed structure into flexible, learnable components. Specifically, we introduce two primary generalizations:

\begin{enumerate}
    \item \textbf{Per-Edge Weighting ($\lambda^{p \cdot q} \to \lambda_{p,q}$):} Instead of the fixed power term $\lambda^{p \cdot q}$ derived from the theorem, we assign a unique, learnable weight $\lambda_{p,q}$ to every edge connecting input dimension $p$ to hidden node $q$.
    \item \textbf{Independent Translation ($\epsilon \cdot q \to \epsilon_q$):} We replace the fixed shift term (typically scaled by index $q$) with independent learnable parameters $\epsilon_q$, allowing for free shifts of the layer functions.
\end{enumerate}

\begin{figure}[t]
    \centering
    \vspace{-0.2cm}
    \begin{tikzpicture}[
        scale=0.80, transform shape,
        node distance=3cm,
        neuron/.style={circle, draw=black, thick, minimum size=1.2cm, align=center, inner sep=0pt},
        input/.style={circle, draw=black, dashed, minimum size=0.8cm, fill=gray!10},
        func/.style={rectangle, draw=blue!80, fill=blue!5, thick, rounded corners, minimum height=0.6cm, minimum width=1.5cm, align=center, font=\small},
        gsfunc/.style={rectangle, draw=red!80, fill=red!5, thick, rounded corners, minimum height=0.6cm, minimum width=2.4cm, align=center, font=\small},
        basis/.style={rectangle, draw=red!80, dashed, fill=white, thick, rounded corners, align=center, font=\footnotesize\bfseries},
        arrow/.style={-Stealth, thick}
    ]

    
    \node[neuron, label={above:\Large $y_q$}] (sum_kan) at (0,0) {\Huge $\Sigma$};
    \node[below=0.8cm of sum_kan, font=\bfseries] {Standard KAN};
    
    \node[input] (x1) at (-4, 1.5) {$x_1$};
    \node[input] (xp) at (-4, 0) {$x_p$};
    \node[input] (xn) at (-4, -1.5) {$x_n$};
    
    \draw[arrow] (x1) -- node[midway, above, func] (f1) {$\phi_{q,1}(x_1)$} (sum_kan);
    \draw[arrow] (xp) -- node[midway, above, func] (fp) {$\phi_{q,p}(x_p)$} (sum_kan);
    \draw[arrow] (xn) -- node[midway, above, func] (fn) {$\phi_{q,n}(x_n)$} (sum_kan);
    
    \node[below=2.2cm of sum_kan, align=center, font=\footnotesize] {
        \textbf{Unique Functions per Edge} \\ 
    };

    
    \begin{scope}[xshift=9cm]
    
        \node[neuron, label={above:\Large $y_q$}] (sum_gs) at (0,0) {\Huge $\Sigma$};
        \node[below=0.8cm of sum_gs, font=\bfseries] {GS-KAN};
        
        \node[input] (gx1) at (-5, 1.5) {$x_1$};
        \node[input] (gxp) at (-5, 0) {$x_p$};
        \node[input] (gxn) at (-5, -1.5) {$x_n$};
        
        \draw[arrow] (gx1) -- node[midway, above, gsfunc] (op1) {
            $\lambda_{q,1} \cdot \mathbf{\psi}(x_1 + \epsilon_q)$
        } (sum_gs);
        
        \draw[arrow] (gxp) -- node[midway, above, gsfunc] (opp) {
            $\mathbf{\lambda_{q,p}} \cdot \mathbf{\psi}(x_p + \mathbf{\epsilon_q})$
        } (sum_gs);
        
        \draw[arrow] (gxn) -- node[midway, above, gsfunc] (opn) {
            $\lambda_{q,n} \cdot \mathbf{\psi}(x_n + \epsilon_q)$
        } (sum_gs);
        
        \node[basis] (master) at (-2.5, 2.6) {Shared Basis $\mathbf{\psi}$};
        
        \draw[->, red!80, dashed, thick] (master) -- (op1.north);

        \node[below=2.2cm of sum_gs, align=center, font=\footnotesize] {
            \textbf{Shared Basis + Linear Transform} \\ 
        };
        
    \end{scope}

    \end{tikzpicture}
    \caption{\textbf{Edge-Node Computation Comparison.} \textbf{Left:} Standard KAN learns a unique function $\phi_{q,p}$ for every connection. \textbf{Right:} GS-KAN learns a single shared basis $\psi$ and adapts it to each edge via learnable scalars $\lambda$ and shifts $\epsilon$. The node $y_q$ aggregates the results.}
    \label{fig:architecture}
    \vspace{-0.7cm}
\end{figure}
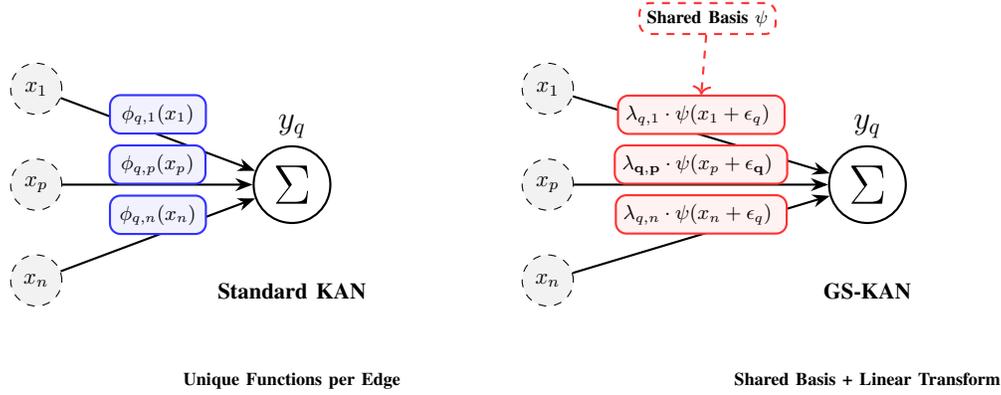

This flexibility allows the network to dynamically discover the optimal separation and alignment of terms in the high-dimensional feature space. Formally, we define the mapping between an input layer of width $N_{in}$ and an output layer of width $N_{out}$ as follows. Let $p \in \{1, \dots, N_{in}\}$ denote the index of the sending node, and $q \in \{1, \dots, N_{out}\}$ denote the receiving node. The value of node $y_q$ is given by:
\begin{equation}
    y_q = \sum_{p=1}^{N_{in}} \lambda_{p,q} \cdot \psi_l \left( x_p + \epsilon_q \right)
\end{equation}

where $\lambda_{p,q}$ represents a learnable edge weight and $\epsilon_q$ is a learnable node-specific bias (translation). See Figure \ref{fig:architecture} for an illustration of this construction.

\paragraph{Distinction from MLPs.} 
It is crucial to distinguish this topology from the standard Multi-Layer Perceptron (MLP). MLPs rely on the Universal Approximation Theorem by applying a fixed non-linearity $\sigma$ \textit{after} a linear summation: $y_q = \sigma(\sum w_{qp} x_p + b_q)$. In contrast, GS-KAN adheres to the Kolmogorov-Arnold topology where learnable non-linearities are applied \textit{on the edges} (pre-summation). While KANs use unique functions $\phi_{qp}$ for every edge, GS-KAN constrains these to be linear transformations of a single shared basis $\psi_l$. This maintains the KAN topology while drastically reducing parameter count.

\subsection{Network Architecture \& Implementation}
We propose a deep architecture constructed by stacking the generalized Sprecher layers described above.

\paragraph{Shared Learnable B-Splines.} 
For each layer $l$, we parameterize the shared basis function $\psi_l(t)$ using (cubic) B-splines. B-splines are fully learnable, allowing the network to discover optimal activation shapes from data. The spline is defined by a set of learnable coefficients $c$ and a knot vector $K$.
\paragraph{Fixed-Domain Strategy.} While it is typically possible for KAN architectures to employ adaptive knot vectors or grid points to track shifting feature distributions, this mechanism inherently introduces additional parameters and computational overhead. Due to parameter efficiency and to maintain a streamlined architecture, we intentionally adopt a standard static domain strategy.  We fix the definition domain of the B-splines to a static interval $[-G, G]$ and the corresponding knot vector to be uniform in this interval. Out-of-bound activations are clamped to the constant value of the nearest boundary knot. Although this yields zero local gradients for these specific samples, batch optimization ensures the aggregate gradient remains informative, allowing the network to dynamically transform activations back into the valid range. Additionally, initializing a slightly wider base domain accommodates intermediate latent representations that may temporarily drift beyond input data boundaries in deeper layers.

\subsection{Parameter Complexity}
The primary motivation for GS-KAN is parameter efficiency. Consider a layer with $N_{in}$ inputs and $N_{out}$ outputs. Let $C$ be the number of parameters per learned function (e.g., B-splines coefficients). For simplicity, we omit linear bias/shift terms ($+N_{out}$) for both MLP and GS-KAN in this comparison.

\begin{itemize}
    \item \textbf{Standard KAN:} Requires a unique function for every edge. 
    Parameter count $\approx N_{in} \times N_{out} \times C$.
    
    \item \textbf{MLP:} Requires one weight per edge. 
    Parameter count $\approx N_{in} \times N_{out}$.
    
    \item \textbf{GS-KAN:} Requires one weight matrix ($\lambda$) and one shared spline ($\psi$). 
    Parameter count $\approx N_{in} \times N_{out} + C$.
\end{itemize}

Since $C$ typically is negligible compared to the weight matrix ($N_{in} \times N_{out}$), GS-KAN achieves the same asymptotic parameter complexity as an MLP ($O(N_{in} \cdot N_{out})$), while retaining the mathematical properties of Kolmogorov-Arnold networks. This contrasts with Standard KANs, which are fundamentally heavier ($O(C \cdot N_{in} \cdot N_{out})$).

\section{Experiments}
We evaluate the performance and parameter efficiency of GS-KAN across three distinct domains: synthetic function approximation, real-world tabular regression, and high-dimensional image classification.

\paragraph{Baselines \& Implementation.} 
We compare GS-KAN against three primary baselines to assess its relative efficiency:
\begin{itemize}
    \item \textbf{MLP:} A standard fully connected Multi-Layer Perceptron. We use {SiLU} activations for function approximation tasks (to favor smoothness) and {ReLU} for tabular/image tasks.
    \item \textbf{Std-KAN:} The standard Kolmogorov-Arnold Network, evaluated using the optimized \texttt{efficient-kan} library \cite{blealtan2024efficient}. Using static unifrom grids for fair comparability.  
    \item \textbf{Wav-KAN:} A wavelet-based KAN implementation using the Mexican Hat wavelet.
\end{itemize}

\paragraph{Training Protocol.} 
All models are implemented in PyTorch and trained using the Adam optimizer, paired with a learning rate scheduler to mitigate training oscillations and ensure stable convergence. Cubic B-Splines were utilized in the KAN and GS-KAN models. To ensure statistical robustness, every experiment is repeated across distinct initialization seeds. To prevent test-set leakage, model selection and early stopping are guided solely by validation set performance. We report the final {test metric} (MSE or Accuracy) by evaluating the optimal validation checkpoint on the held-out test set exactly once, thereby providing a robust measure of true generalization.

\subsection{Synthetic Function Approximation} To evaluate the capability of GS-KAN to model complex, high-frequency, and non-linear interactions under a strict parameter budget ($\approx$ 200 parameters), we conducted a regression benchmark across five diverse synthetic functions.\paragraph{Experimental Setup.}The benchmark comprises three 2D and two 3D target functions, detailed in Table \ref{tab:synth_funcs}. The input variables were uniformly sampled from the domain $[-1, 1]^d$, where $d \in \{2, 3\}$. To simulate realistic, noisy conditions, additive Gaussian noise $\epsilon \sim \mathcal{N}(0, 0.01^2)$ was introduced to all targets, establishing a theoretical lower bound for the Mean Squared Error (MSE) of $1.0 \times 10^{-4}$.

\begin{table}[htpb]
\centering
\vspace{-0.7cm}
\caption{Definitions of the 2D and 3D synthetic target functions.}
\label{tab:synth_funcs}
\renewcommand{\arraystretch}{1.2} 
\scalebox{0.85}{
\begin{tabular}{|c|l|l|}
\hline
\textbf{ID} & \textbf{Type (Dimension)} & \textbf{Function Definition} \\
\hline
F1 & High-Freq Ripple (2D) & $f(x) = \sin(3\pi x_1) \cos(3\pi x_2)$ \\
\hline
F2 & Bessel Function (2D) & $f(x) = J_0\left(20\sqrt{x_1^2 + x_2^2}\right)$ \\
\hline
F3 & Polynomial (2D) & $f(x) = x_1^2 x_2 - 3x_1 x_2^3 + x_1 x_2 + 2x_1^3 x_2^2$ \\
\hline
F4 & Exp-Trig (3D) & $f(x) = \exp(x_1 x_2 + \cos(3x_3)) \sin(2x_1)$ \\
\hline
F5 & Rational (3D) & $f(x) = \frac{x_2 + x_3}{1 + x_1^2} + \frac{x_2^2 - x_1^3}{2 - x_2 x_3 + x_3^2 + x_1}$ \\
\hline
\end{tabular}}
\vspace{-0.7cm}
\end{table}

\paragraph{Training Details.} For each function, a dataset of 4,096 samples was generated and partitioned into training (70\%), validation (15\%), and test (15\%) sets. The models were trained for 150 epochs with a batch size of 128, utilizing the Adam optimizer with an initial learning rate of 0.01 modulated by an exponential decay scheduler ($\gamma = 0.99$). The experiment was averaged over 10 independent random seeds. Empirically, the operational domains were optimized for each architecture. While standard KAN achieved its best results within the $[-1, 1]$ range, gaining no advantage from domain expansion, GS-KAN required a wider domain of $[-2, 2]$ for optimal performance. We hypothesize that GS-KAN's deeper latent architecture benefits from this wider spatial separation to more effectively distribute intermediate representations.

\begin{figure}[htbp]
    \centering
    \vspace{-0.0cm}
    \includegraphics[width=0.85\linewidth]{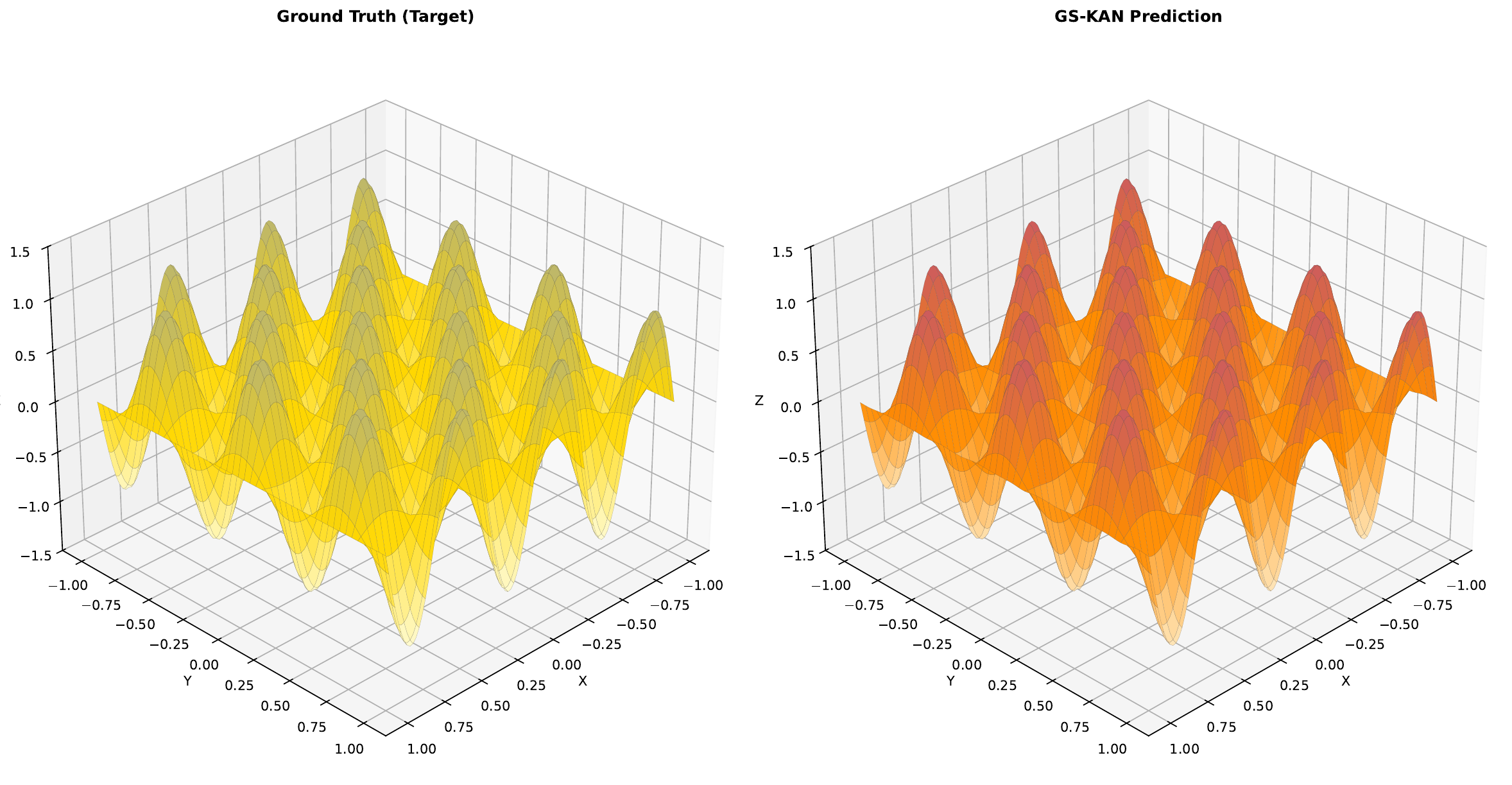}
    \caption{\textbf{Function Approximation Visualization.} 
    \textbf{Left:} The ground truth function $f(x,y) = \sin(3\pi x) \cdot \cos(3\pi y)$ (noiseless). 
    \textbf{Right:} The approximation generated by GS-KAN ($\approx 200$ parameters). 
   }
    \label{fig:gskan_surface}
    \vspace{-0.2cm}
\end{figure}

\paragraph{Results.} The test mean squared error (MSE) results for these experiments are presented in Table \ref{tab:function_experiments} and an illustration of a GS-KAN prediction for $F_1$ is found in Figure \ref{fig:gskan_surface}.

\begin{table}[t]
\centering
\caption{Test MSE (Mean $\pm$ Std) for function approximation tasks. All MSE values are scaled by $10^3$ (i.e., reported as $\times 10^{-3}$) for readability. The best performing model (lowest mean error) for each task is highlighted in bold.}
\label{tab:function_experiments}
\resizebox{0.8\textwidth}{!}{
\begin{tabular}{l l c | c | c | c}
\toprule
\multicolumn{6}{c}{\textbf{2D Input Functions}} \\
\midrule
\textbf{Model} & \textbf{Structure} & \textbf{Params} & \textbf{F1: Ripple} & \textbf{F2: Bessel} & \textbf{F3: Polynomial} \\
\midrule
GS-KAN  & [2, 10, 9, 1]  & 190 & $\mathbf{0.74} \pm \mathbf{1.14}$ & $6.84 \pm 3.53$ & $0.27 \pm 0.10$ \\
Wav-KAN & [2, 7, 7, 1]   & 210 & $33.00 \pm 13.70$ & $22.20 \pm 4.55$ & $0.80 \pm 0.85$ \\
MLP     & [2, 12, 12, 1] & 205 & $252.00 \pm 5.87$ & $28.30 \pm 2.68$ & $0.48 \pm 0.16$ \\
Std-KAN & [2, 5, 1]      & 195 & $1.46 \pm 0.64$ & $\mathbf{6.45} \pm \mathbf{1.95}$ & $\mathbf{0.13} \pm \mathbf{0.01}$ \\
\midrule
\midrule
\multicolumn{6}{c}{\textbf{3D Input Functions}} \\
\midrule
\textbf{Model} & \textbf{Structure} & \textbf{Params} & \textbf{F4: Exp-Trig} & \textbf{F5: Rational} & \textbf{-} \\
\midrule
GS-KAN  & [3, 9, 9, 1]   & 187 & $\mathbf{1.01} \pm \mathbf{0.76}$ & $\mathbf{0.41} \pm \mathbf{0.10}$ & - \\
Wav-KAN & [3, 7, 6, 1]   & 207 & $3.04 \pm 1.21$ & $1.18 \pm 0.37$ & - \\
MLP     & [3, 12, 11, 1] & 203 & $1.28 \pm 0.31$ & $0.80 \pm 0.30$ & - \\
Std-KAN & [3, 4, 1]      & 208 & $5.60 \pm 5.68$ & $0.60 \pm 0.32$ & - \\
\bottomrule
\end{tabular}
}
\vspace{-0.4cm}
\end{table}
\footnotetext{For GS-KAN, the number of coefficients for each function is $N_c = K - d - 1$, whereas for Standard KAN, it is $N_c = G + d$, where $d=3$ is the spline degree.}

\paragraph{Analysis.}
Despite utilizing the fewest learnable parameters, GS-KAN performed the best overall out of all the evaluated baselines. Specifically, GS-KAN achieved the lowest mean squared error for F1, F4, and F5, while having a slightly higher error than Std-KAN on the Bessel and Polynomial tasks (F2 and F3). These results are particularly promising considering that mathematical function approximation is an area where standard KAN networks are traditionally known to excel. Std-KAN demonstrates a slight advantage on smoother, highly structured 2D functions (F2 and F3). However, GS-KAN proves more capable of capturing complex, higher-dimensional, or high-frequency mappings (F1, F4, and F5). Furthermore, the baseline models MLP and Wav-KAN exhibit inconsistency across the tests, for instance they have order of magnitude higher error on F1 and F2 compared to the other two. In contrast, GS-KAN and Std-KAN are robust, successfully approximating all five diverse functions without catastrophic failures.

\subsection{Real-World Non-Linear Regression (kin8nm)}To evaluate the models on real-world, non-linear relationships, we conducted a regression experiment using the kin8nm dataset \cite{kin8nm}. This task involves predicting the forward kinematics of an 8-link robot arm, providing a highly non-linear mapping challenge.\paragraph{Experimental Setup.}The dataset consists of 8,192 observations. The 8 input features represent joint angles bounded between $-\pi/2$ and $\pi/2$, centered around zero with a variance of approximately $0.9$. Because the raw data is naturally bounded and zero-centered, it is highly compatible with the static grid intervals used in the tested KAN architectures. Consequently, no prior feature scaling or standardization was applied. All experiments were averaged over 10 independent random seeds. For each seed, the data was randomly partitioned into training (70\%), validation (15\%), and test (15\%) sets. Models were evaluated across three parameter budgets: $\approx 200$, $\approx 600$, and $\approx 2000$ parameters.\paragraph{Training Details.} All models were trained for 100 epochs using a batch size of 128 to minimize the Mean Squared Error (MSE). We used the Adam optimizer with an initial learning rate of 0.01. To facilitate convergence to a precise minimum, a \textit{ReduceLROnPlateau} scheduler was applied. For both GS-KAN and Standard KAN, the B-spline grid domain was set to $[-3, 3]$. Although the raw input features are bounded within $[-\pi/2, \pi/2]$, empirical evaluation demonstrated that a wider grid yielded superior performance. While a $[-3, 3]$ domain slightly underutilizes the grid capacity in the first input layer, it provides spatial headroom for the latent representations in deeper layers, where intermediate activations may be transformed beyond the original input range.

\paragraph{Results.} 
We benchmarked performance across three parameter regimes. Results are shown in Table \ref{tab:kin8nm_results}.

\begin{table}[htbp]
\centering
\vspace{-0.4cm}
\caption{\textbf{Robot Arm Kinematics (kin8nm) Results.} Comparison of Best Test MSE (mean $\pm$ std) . The best result in each budget regime is highlighted in \textbf{bold}. Values are scaled by $10^{3}$ (i.e., reported as $\times 10^{-3}$) for readability.}
\label{tab:kin8nm_results}
\vspace{0.1cm}
\footnotesize
\setlength{\tabcolsep}{4pt} 
\scalebox{0.85}{
\begin{tabular}{@{}llclcc@{}}
\toprule
\textbf{Model} & \textbf{Architecture} & \textbf{Res.} & \textbf{Params} & \textbf{Best Test MSE} \\ 
\midrule
\multicolumn{5}{l}{\textit{\textbf{Budget Regime I} ($\approx 200$ Params)}} \\ 
\midrule 
MLP & $[8, 10, 10, 1]$ & - & 211 & $8.39 \pm 0.48$ \\
Std-KAN & $[8, 2, 1]$ & $G=7$ & 216 & $20.90 \pm 0.90$ \\
Wav-KAN & $[8, 5, 5, 1]$ & - & 210 & $20.32 \pm 8.11$ \\
GS-KAN & $[8, 8, 8, 1]$ & $K=20$ & 204 & $\mathbf{6.43 \pm 0.44}$ \\
\midrule
\multicolumn{5}{l}{\textit{\textbf{Budget Regime II} ($\approx 600$ Params)}} \\ 
\midrule 
MLP & $[8, 20, 20, 1]$ & - & 621 & $6.12 \pm 0.47$ \\
Std-KAN & $[8, 5, 1]$ & $G=8$ & 585 & $8.55 \pm 0.67$ \\
Wav-KAN & $[8, 11, 10, 1]$ & - & 624 & $31.01 \pm 15.96$ \\
GS-KAN & $[8, 18, 18, 1]$ & $K=20$ & 574 & $\mathbf{4.48 \pm 0.20}$ \\
\midrule
\multicolumn{5}{l}{\textit{\textbf{Budget Regime III} ($\approx 2000$ Params)}} \\ 
\midrule 
MLP & $[8, 40, 40, 1]$ & - & 2041 & $5.25 \pm 0.24$ \\
Std-KAN & $[8, 8, 8, 1]$ & $G=10$ & 2040 & $\mathbf{4.80 \pm 0.28}$ \\
Wav-KAN & $[8, 22, 21, 1]$ & - & 1977 & $40.67 \pm 5.85$ \\
GS-KAN & $[8, 38, 38, 1]$ & $K=40$ & 1974 & $\mathbf{4.80 \pm 0.29}$ \\
\bottomrule
\end{tabular}}
\vspace{-0.7cm}
\end{table}

\paragraph{Analysis.} As shown in Table \ref{tab:kin8nm_results}, GS-KAN is the best performing model. It consistently outperforms the baseline MLP and other KAN variants in the highly constrained parameter regimes ($\approx 200$ and $\approx 600$ parameters). Interestingly, while Wav-KAN performed adequately on synthetic tasks, its error increases progressively on this dataset, suggesting it may have limited versatility across different data distributions. Notably, GS-KAN achieves its lowest MSE ($4.48 \times 10^{-3}$) within the $\approx 600$ parameter regime. However, increasing the capacity to $\approx 2000$ parameters leads to a slight degradation in performance ($4.80 \times 10^{-3}$), which could suggest poor scaling and indicate a tendency to overfit this specific dataset. Nevertheless, the best-performing architectures ultimately concentrate around a similar MSE of approximately $5 \times 10^{-3}$ in the largest budget regime. This may indicate that the inherent noise floor or irreducible error of the kin8nm dataset lies near this threshold. Consequently, GS-KAN may have reached the dataset's representational limit at a fraction of the capacity required by the baselines.

\subsection{High-Dimensional Classification (Fashion-MNIST)}To assess the versatility of GS-KAN beyond regression tasks, we evaluated its performance on a high-dimensional classification problem using the Fashion-MNIST dataset \cite{xiao2017fashion}. This task tests the models' ability to handle large input vectors ($D_{in}=784$) and extract semantic features under strict parameter constraints, without relying on the inductive bias of convolutional layers.\paragraph{Experimental Setup.}The Fashion-MNIST dataset comprises 70,000 grayscale images distributed across 10 classes. The data was partitioned into 50,000 training, 10,000 validation, and 10,000 test samples. All images were flattened into 784-dimensional vectors and normalized to the $[-1, 1]$ range. To rigorously test representation efficiency, all evaluated models were restricted to a tight parameter budget of approximately 10k parameters.\paragraph{Training Details.}Models were trained for 20 epochs with a batch size of 256, optimizing the Cross-Entropy loss. We utilized the Adam optimizer with an initial learning rate of 0.001, modulated by an exponential decay scheduler ($\gamma = 0.95$). For GS-KAN, the B-spline grid domain was set to $[-3, 3]$ and  standard KAN grid was kept at $[-1, 1]$, as the vast majority of its splines act directly on the normalized inputs, and extending its grid yielded no empirical benefits. The experiment was averaged over 5 independent random seeds.

\paragraph{Results.} 
We compare GS-KAN against the baseline models. To ensure a rigorous comparison, the MLP was configured with a slightly larger parameter budget than the GS-KAN model.

\begin{table}[htbp]
\centering
\vspace{-0.0cm} 
\caption{Fashion-MNIST test accuracy (mean $\pm$ std). Best results in \textbf{bold}.}
\label{tab:fashion_mnist_results}
\scalebox{0.85}{
\begin{tabular}{llcc}
\toprule
\textbf{Model} & \textbf{Structure} & \textbf{Params} & \textbf{Accuracy (\%)} \\
\midrule
GS-KAN  & [784, 12, 12, 10] & 9877  & $\mathbf{85.59 \pm 0.33}$ \\
MLP     & [784, 13, 12, 10] & 10503 & $84.87 \pm 0.26$ \\
Std-KAN & [784, 1, 10]      & 10322 & $60.01 \pm 1.56$ \\
Wav-KAN & [784, 4, 10]      & 9528  & $10.20 \pm 0.98$ \\
\bottomrule
\end{tabular}}
\vspace{-0.7cm} 
\end{table}

\paragraph{Analysis.}
The results illustrate the parameter explosion problem inherent in standard KAN architectures. To adhere to the strict 10k parameter constraint, both Std-KAN and Wav-KAN had to be restricted to critically narrow latent layers. This bottleneck resulted in poor learning for Std-KAN and a complete failure to capture structural patterns for Wav-KAN, which effectively degenerated into random guessing. In contrast, despite utilizing fewer parameters, GS-KAN outperforms the MLP baseline by an average margin of $0.72\%$. 

\section{Conclusion and Future Work}
In this work, we presented GS-KAN, a parameter-efficient adaptation of Kolmogorov-Arnold Networks inspired by Sprecher's refinement of the superposition theorem. By utilizing layer-wise shared learnable B-splines and linear transformations, GS-KAN effectively decouples network width from the parameter complexity of the learned functions. Our results demonstrate that this architecture offers a highly expressive and versatile alternative to existing models.

Unlike earlier numerical implementations of ``Sprecher networks''—which necessitated pathological or fractal-like inner functions to strictly adhere to the theorem—our relaxation distributes the informational burden across generalized coefficients. This structural flexibility suggests that smooth B-spline approximations are sufficient for high-precision learning within this framework, bridging the gap between exact representation theory and practical gradient-based optimization. Under strict parameter constraints, GS-KAN consistently outperformed the evaluated baselines across synthetic function approximation, real-world tabular regression, and high-dimensional image classification. This suggests that the layer-wise shared basis strategy is a promising architectural direction, enabling spline-based learning to scale efficiently to a wider range of general machine learning applications.

\paragraph{Limitations and Future Work:}

While the preliminary validation of the GS-KAN architecture yields promising results on constrained memory budgets, this work represents an initial exploratory study. We identify the following current limitations, which outline important directions for future research:

\begin{itemize}
    \item \textbf{Computational Overhead and Training Time:} Empirically, the current implementation of GS-KAN requires approximately 2x to 10x longer time to train compared to optimized KAN variants, depending on network size. We attribute this latency to an engineering bottleneck rather than a fundamental theoretical limitation. Our current code base does not yet incorporate the advanced memory and computational optimizations found in frameworks such as Efficient KAN \cite{blealtan2024efficient}. Because GS-KAN inherently relies on a reduced number of trainable B-spline parameters, we hypothesize that a fully optimized implementation could eventually match or exceed the training speed of optimized standard KANs.
    
    \item \textbf{Adaptive Grid Distributions:} Because each layer in GS-KAN relies on a single shared function to process all inputs, maximizing the expressive efficiency of this spline is critical. Currently, the model employs fixed, uniform spline grids, which can inefficiently allocate knots to regions with low data density or restrict latent representations to suboptimal values. Implementing data-driven learnable knot positions would allow the architecture to dynamically concentrate grid resolution on the most complex and dense regions of the input space, thereby enhancing both parameter efficiency and overall performance. It would also make the B-spline domain selection autonomous. 

    \item \textbf{Scaling Behavior:} The empirical evaluations in this study are currently limited to smaller-scale architectures and datasets. Further investigation is required to understand the scaling laws of GS-KAN. It remains to be seen how the network's performance, stability, and representational capacity evolve when scaling up to deeper networks and higher-dimensional, complex datasets. Dynamic grid optimization with an ability to extend grid resolution could be important in such a study. 
\end{itemize}

\section*{Acknowledgements}
The author would like to thank S.Zuyev, J. Larsen, A. Malmquist, and M. Redin for valuable discussions and feedback during the early conception of this work.


\end{document}